\documentclass[12pt,a4paper]{cibb}

\makeatletter
\providecommand{\@ordinalM}[2]{#1}
\makeatother

\usepackage{subfigure,graphicx}
\usepackage{amsmath,amsfonts,latexsym,amssymb,euscript,xr}
\usepackage{booktabs}
\usepackage[nodayofweek]{datetime}
\usepackage{hyperref}
\usepackage{fmtcount}
\usepackage[english]{datenumber}
\usepackage[absolute]{textpos}
\usepackage[table]{xcolor}
\usepackage{color,colortbl,tabularx}
\usepackage[english]{babel}
\usepackage[protrusion=true,expansion=true]{microtype}
\usepackage{amsthm}
\usepackage{pifont}
\usepackage{placeins}
\usepackage{float}

\makeatletter
\let\origthebibliography\thebibliography
\let\endorigthebibliography\endthebibliography
\renewenvironment{thebibliography}[1]{%
  \origthebibliography{#1}%
  \footnotesize
  \setlength{\itemsep}{-2pt}%
  \setlength{\parskip}{0pt}%
  \setlength{\parsep}{0pt}%
  \linespread{0.92}\selectfont
}{\endorigthebibliography}
\makeatother

\definecolor{LightBlue}{rgb}{0.88,0.9,0.9}

\title{\Large $\ $\\ \bf Reliability-Aware Benchmarking of
Self-Supervised Learning in Histopathology
Under 1--10\% Label Regimes}

\author{\large Alicem Koyun$^{*,1}$, Pranjali Changadeo Pingale$^{1}$, Sachin Mote$^{2}$, and Gianluca Pizzuti$^{3}$}
\address{\footnotesize $\ $\\$^1$ Westcliff University, Irvine, CA, United States. \\
$^2$ George Mason University, Fairfax, VA, United States. \\
$^3$ University of Miami, Coral Gables, FL, United States. \\
\bigskip
ORCID codes: Alicem Koyun 0009-0005-5152-3500; Pranjali Changadeo Pingale 0009-0004-5377-2351; Sachin Mote 0009-0004-7082-7844; Gianluca Pizzuti 0009-0000-6274-1216.
\bigskip
\newline
$^*$corresponding author: alicemkoyun@gmail.com

}

\abstract{\small
self-supervised learning, histopathology, label efficiency, calibration, federated learning.
\normalsize \\[17pt]
{\bf Abstract.}
Expert annotation in digital pathology is expensive, and low-label regimes remain common in realistic histopathology workflows. Self-supervised learning (SSL) offers a practical way to learn transferable representations from unlabeled pathology images, but performance-only evaluation can hide unreliable confidence behavior under domain shift. We present a reliability-aware benchmark of SimCLR, BYOL, and an MAE-inspired masked-reconstruction baseline on PatchCamelyon (PCam; 96$\times$96) and PANDA (prostate cancer) under 1\%, 5\%, and 10\% labeled regimes with frozen-encoder linear probing. To address common single-run concerns, ResNet-18 results are repeated over 3 splits $\times$ 3 seeds ($n=9$) and ViT-B/16 over 1 split $\times$ 3 seeds ($n=3$) under practical compute constraints, with mean$\pm$std reporting and per-run logs. Beyond AUROC, accuracy, and F1-score, we report Expected Calibration Error and Brier score, OOD confidence diagnostics for cross-dataset PCam$\leftrightarrow$PANDA scoring, federated-vs-centralized SimCLR comparisons on PCam, and cross-dataset transfer to colorectal histology tiles. SSL improves label efficiency on the more controlled PCam task, while PANDA's noisier patch-level proxy can remain near coin-flip with higher variance. The study highlights that histopathology SSL should be assessed not only by ranking performance but also by calibration, robustness, and transparent reproducibility diagnostics before clinical interpretation.}

\begin{document}

\renewcommand{\thefootnote}{}
\footnotetext{\small{Article version: 5th July 2026 $\;$ h08:53 $\;$ CET}}

\thispagestyle{myheadings}
\pagestyle{myheadings}
\markright{\tt Proceedings of CIBB 2026}

\section{Introduction}
\label{sec:intro}

Digital pathology has rapidly advanced with high-resolution whole-slide imaging and deep learning, enabling large-scale analysis of histological patterns for cancer detection, grading, and biomarker discovery~\cite{wang2022precisiononcology}. However, state-of-the-art systems rely heavily on expert-labeled data, and the fine-grained nature of histological patterns makes WSI annotation slow and costly, creating a significant bottleneck for clinical-grade models~\cite{campanella2019clinical}.

Self-supervised learning (SSL) reduces dependence on exhaustive annotations by optimizing pretext tasks, such as contrastive or reconstruction-based objectives, on unlabeled data and transferring to downstream tasks with limited labels~\cite{grill2020byol,chen2020simclr}. This is particularly appealing in histopathology, where large archives of unlabeled WSIs are available while fully annotated datasets remain scarce, yet relatively few works systematically compare SSL frameworks under low-label regimes with reliability metrics.

Despite rapid adoption of SSL in digital pathology, a basic practical question remains: with only 1--10\% labels, which SSL paradigm yields the most useful and reliable model? Many prior pathology SSL studies report a single split/seed, focus on AUROC, and omit calibration or robustness checks, leaving it unclear whether label-efficiency gains are stable and whether high-AUROC representations stay trustworthy under shift.

Rather than introducing another SSL objective, this paper focuses on a controlled and reliability-aware evaluation protocol for existing SSL paradigms in histopathology, where low-label performance, calibration, and domain-shift behavior are often more clinically relevant than isolated single-run accuracy. We compare SimCLR and BYOL at 1\%, 5\%, and 10\% labels using frozen-encoder linear probing, reporting AUROC/accuracy/F1 together with reliability metrics (ECE, Brier). To address single-run concerns, ResNet-18 results are averaged over 3 splits $\times$ 3 seeds ($n=9$), while ViT-B/16 uses one fixed split with three seeds ($n=3$) due to Kaggle runtime limits.

\textbf{Contributions.} (i) A controlled low-label SSL benchmark for histopathology covering SimCLR, BYOL, and an MAE-inspired baseline at 1\%, 5\%, and 10\% labels. (ii) Reliability-aware evaluation that pairs ranking metrics with ECE, Brier score, and OOD confidence diagnostics. (iii) Reproducibility through fixed splits, repeated seeds, and mean$\pm$std reporting. (iv) Diagnostic extensions via federated SimCLR and cross-dataset transfer to colorectal histology to probe representation behavior under distribution shift.

\section{Data and Methods}
\label{sec:methods}

\subsection{Datasets, task definition, and preprocessing}

\textbf{PatchCamelyon (PCam).} PCam is a binary classification dataset of tumor versus non-tumor breast cancer patches extracted from CAMELYON16/17 WSIs~\cite{veeling2018pcam}. We use the official training split (approximately 220k patches) and treat each patch as an independent sample; this patch-level setting is intentionally simple to isolate representation quality from slide-level aggregation effects.

\textbf{PANDA.} The PANDA dataset comprises prostate cancer WSIs originally curated for Gleason grading~\cite{bulten2022panda}. We adopt a patch-based proxy task by sampling resized $224{\times}224$ tiles from the provided images. To align with binary evaluation, we derive labels from the original grade information (clinically significant vs. non-significant cancer). We emphasize that this binarization is a pragmatic proxy for benchmarking label efficiency rather than a clinically validated endpoint.

\begin{table}[t]
\footnotesize
\renewcommand{\arraystretch}{0.92}
\centering
\begin{tabular}{@{}lccc@{}}
\toprule
\textbf{Split} & \textbf{Train 0/1} & \textbf{Val 0/1} & \textbf{Overlap} \\
\midrule
\rowcolor{LightBlue} 0 & 844 / 854 & 211 / 214 & 0 \\
1 & 844 / 854 & 211 / 214 & 0 \\
\rowcolor{LightBlue} 2 & 844 / 854 & 211 / 214 & 0 \\
\bottomrule
\end{tabular}
\caption{\textbf{PANDA binary proxy split checks.} The experimental subset contains 1055 class-0 and 1068 class-1 samples, with stratified train/validation partitions and zero index overlap.\label{tab:panda_validity}}
\vspace{0.8mm}
\end{table}

\textbf{Colorectal histology.} We use colorectal histology tiles with eight tissue categories for cross-dataset transfer~\cite{kather2016textures}. The transfer experiment evaluates multi-class linear probing on frozen features using a fixed subset of 5,000 tiles for efficiency. This setup probes whether PCam representations preserve broad tissue semantics beyond the tumor/non-tumor axis.

PCam provides $96{\times}96$ RGB patches, while PANDA and colorectal experiments use $224{\times}224$ RGB patches or resized crops. Images are normalized channel-wise with mean and std $(0.5,0.5,0.5)$. During SSL pretraining we apply RandomResizedCrop$(0.6,1.0)$, horizontal flips (vertical flips for PCam), ColorJitter$(0.2,0.2,0.2,0.1)$, and RandomGrayscale$(p=0.1)$.

\subsection{Self-supervised methods and evaluation protocol}

SimCLR maximizes agreement between differently augmented views using an InfoNCE loss~\cite{chen2020simclr}. We attach a two-layer MLP projection head (hidden 2048, output 128) and use temperature $\tau=0.5$. BYOL trains an online network to predict the representation of a momentum-updated target network~\cite{grill2020byol}. We implement online and target networks with an MLP projector/predictor (hidden 4096, projection 256) and EMA decay 0.996. Motivated by masked autoencoding, we also include a lightweight MAE-inspired masked-reconstruction baseline trained with fixed mask ratio 0.6 and block size 32~\cite{he2022mae}.

For PCam, we simulate federated SimCLR with 3 clients performing local SSL training and periodic aggregation via FedAvg~\cite{mcmahan2017fedavg}. Each communication round performs 1 local epoch of SimCLR followed by FedAvg; we run 5 global rounds.

Feature quality is evaluated using a frozen-encoder linear probe at label fractions of 1\%, 5\%, and 10\%. For each label fraction, we train a linear classifier on frozen SSL features for 10 epochs using Adam ($10^{-3}$ learning rate, batch size 128). For ResNet-18, we evaluate 3 predefined split identifiers and 3 random seeds per split ($n=9$). For ViT-B/16, we keep one split fixed and vary 3 seeds ($n=3$) due to compute constraints, while preserving identical splits across methods for fairness.

We report AUROC, accuracy, and F1-score as ranking metrics, and Expected Calibration Error (ECE; 15 equal-width bins) and Brier score as reliability metrics~\cite{guo2017calibration}; reporting both helps distinguish high-AUROC models that are nonetheless overconfident. For OOD detection, we use binary MSP confidence $\max(p,1-p)$ and uncertainty $u=1-\mathrm{MSP}$. AUROC$_{\mathrm{raw}}$ treats OOD as positive using $u$ as the score; AUROC$_{\mathrm{flip}}=1-\mathrm{AUROC}_{\mathrm{raw}}$ is reported only as an inversion diagnostic when the score direction reverses.

\section{Results}
\label{sec:results}

\subsection{Main label-efficiency benchmark}

\begin{figure}[!htbp]
\begin{center}
\begin{minipage}{0.48\textwidth}
  \includegraphics[width=\linewidth]{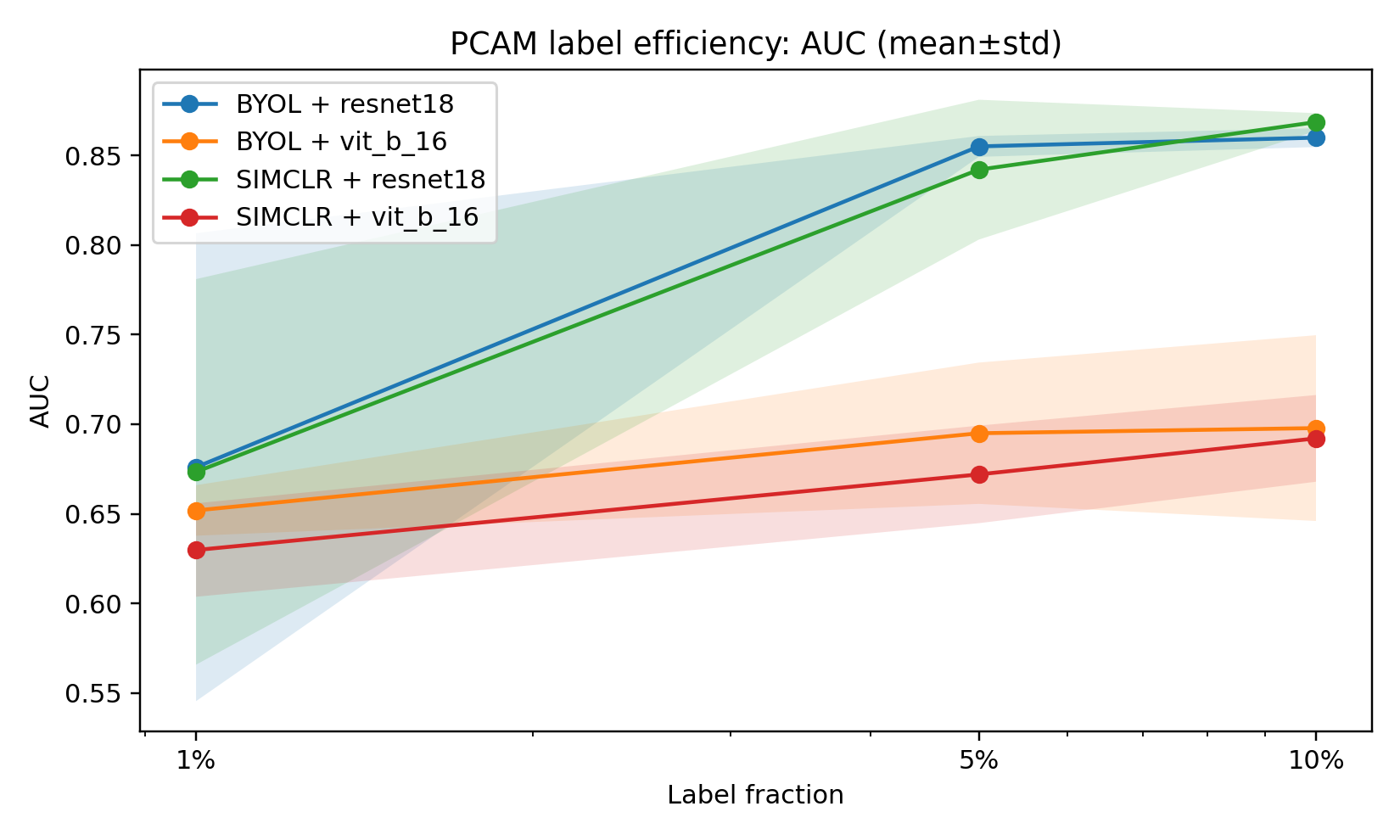}
  \centerline{\small (a) PCam}
\end{minipage}\hfill
\begin{minipage}{0.48\textwidth}
  \includegraphics[width=\linewidth]{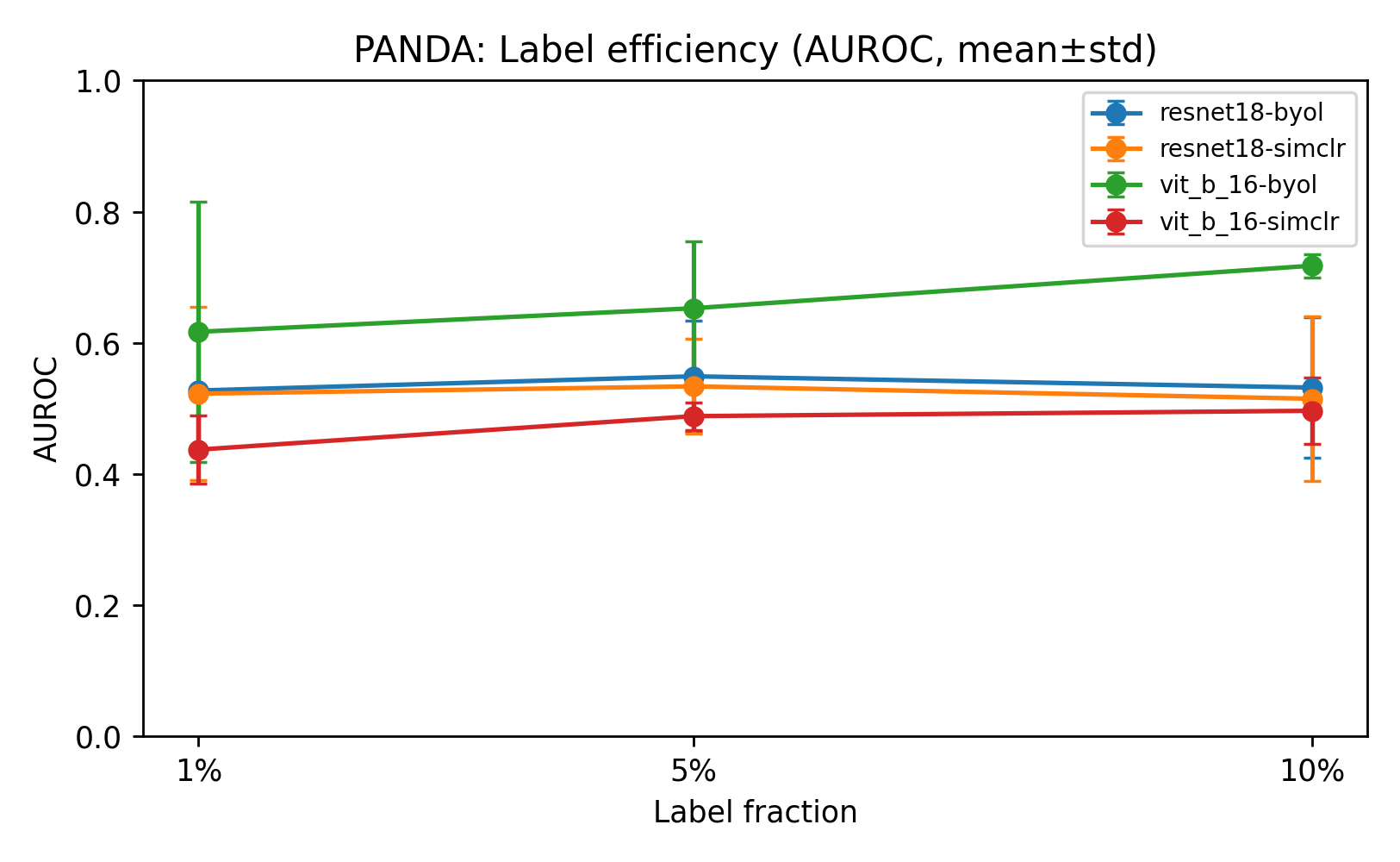}
  \centerline{\small (b) PANDA}
\end{minipage}
\caption{\textbf{Label efficiency under 1--10\% labelled regimes.} AUROC is reported as mean$\pm$std. ResNet-18 uses $n=9$; ViT-B/16 uses $n=3$ with one fixed split.\label{fig:label_efficiency}}
\end{center}
\vspace{-3mm}
\end{figure}

\autoref{fig:label_efficiency} summarizes AUROC versus label fraction for PCam and PANDA. Across both datasets, the mean$\pm$std curves make clear that ViT-B/16+BYOL is more label-efficient on PANDA, while ResNet-18 differences are smaller. Since ViT-B/16 uses one fixed split with three seeds, we read its comparison to the $n=9$ ResNet-18 runs as indicative rather than a matched statistical test. On PCam, SimCLR is competitive at 1\% and 10\%, and BYOL can be stronger at 5\%. Importantly, the variance over runs is non-negligible in low-label settings, motivating repeated reporting rather than single-point estimates.

\begin{table}[t]
\footnotesize
\renewcommand{\arraystretch}{0.92}
\centering
\begin{tabularx}{0.98\textwidth}{lXccc}
\toprule
\textbf{Dataset} & \textbf{Model / method} & \textbf{Labels} & \textbf{AUROC} & \textbf{ECE} \\
\midrule
\rowcolor{LightBlue} PCam & ResNet-18 SimCLR & 1\% & $0.6760\pm0.1060$ & $0.1174\pm0.0157$ \\
PCam & ResNet-18 SimCLR & 10\% & $0.8685\pm0.0049$ & $0.0828\pm0.0079$ \\
\rowcolor{LightBlue} PCam & ResNet-18 BYOL & 10\% & $0.8599\pm0.0053$ & $0.0900\pm0.0075$ \\
PANDA & ResNet-18 BYOL & 1\% & $0.5854\pm0.1011$ & $0.0871\pm0.0486$ \\
\rowcolor{LightBlue} PANDA & ResNet-18 BYOL & 10\% & $0.5239\pm0.0737$ & $0.0699\pm0.0389$ \\
PANDA & ViT-B/16 BYOL & 10\% & $0.7176\pm0.0178$ & $0.0806\pm0.0208$ \\
\bottomrule
\end{tabularx}
\caption{\textbf{Selected low-label and 10\% label results.} A compact view of the most informative quantitative findings; full metric tables are omitted to respect the CIBB page limit.\label{tab:main_summary}}
\vspace{0.4mm}
\end{table}

The PANDA downstream task is intentionally a patch-level proxy derived from slide-level ISUP grades (ISUP$\geq$2 vs. below) on resized tiles. This introduces label noise: many sampled tiles may not contain discriminative morphology for the assigned slide label, and the prostate domain also exhibits substantial appearance variability and class overlap. Under low label budgets and with a small backbone, these factors can yield AUROC values close to 0.5 with large variance. The non-monotonic AUROC trend on PANDA (1\%$\rightarrow$10\%) is consistent with the high run-to-run variance and overlapping confidence intervals reported in \autoref{tab:main_summary}, reflecting noisy proxy supervision rather than a methodological inconsistency. We therefore report near-chance AUROC values on PANDA as a transparent diagnostic outcome rather than hide them.

\textbf{Sanity baselines (supervised).} To confirm the pipeline is learnable under the same splits, we add supervised ResNet-18 baselines; ImageNet initialization improves ranking and calibration over training from scratch on both datasets and anchors SSL gains under identical compute.

\subsection{Reliability and calibration}

\begin{figure}[!htbp]
\begin{center}
\begin{minipage}{0.48\textwidth}
  \includegraphics[width=\linewidth]{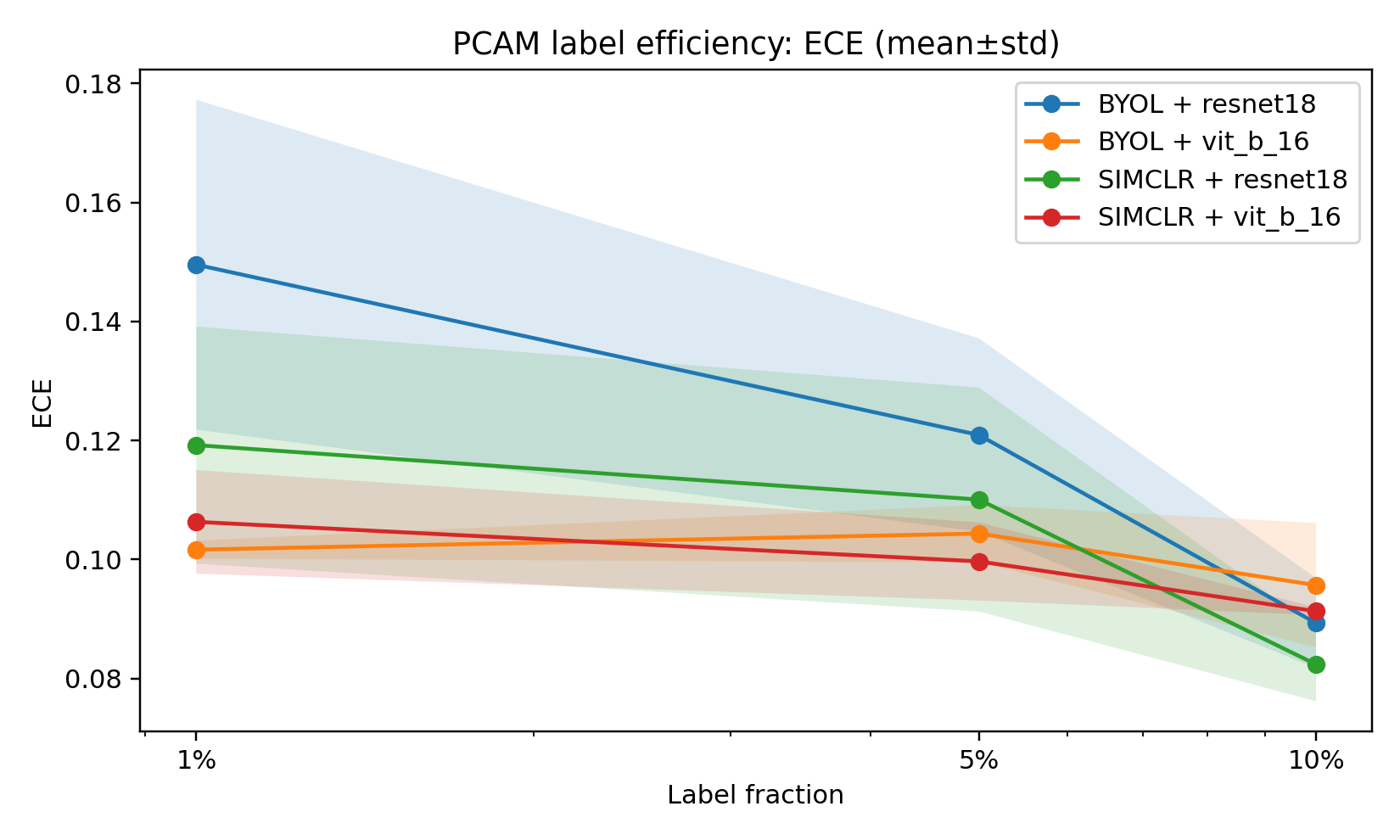}
  \centerline{\small (a) ECE}
\end{minipage}\hfill
\begin{minipage}{0.48\textwidth}
  \includegraphics[width=\linewidth]{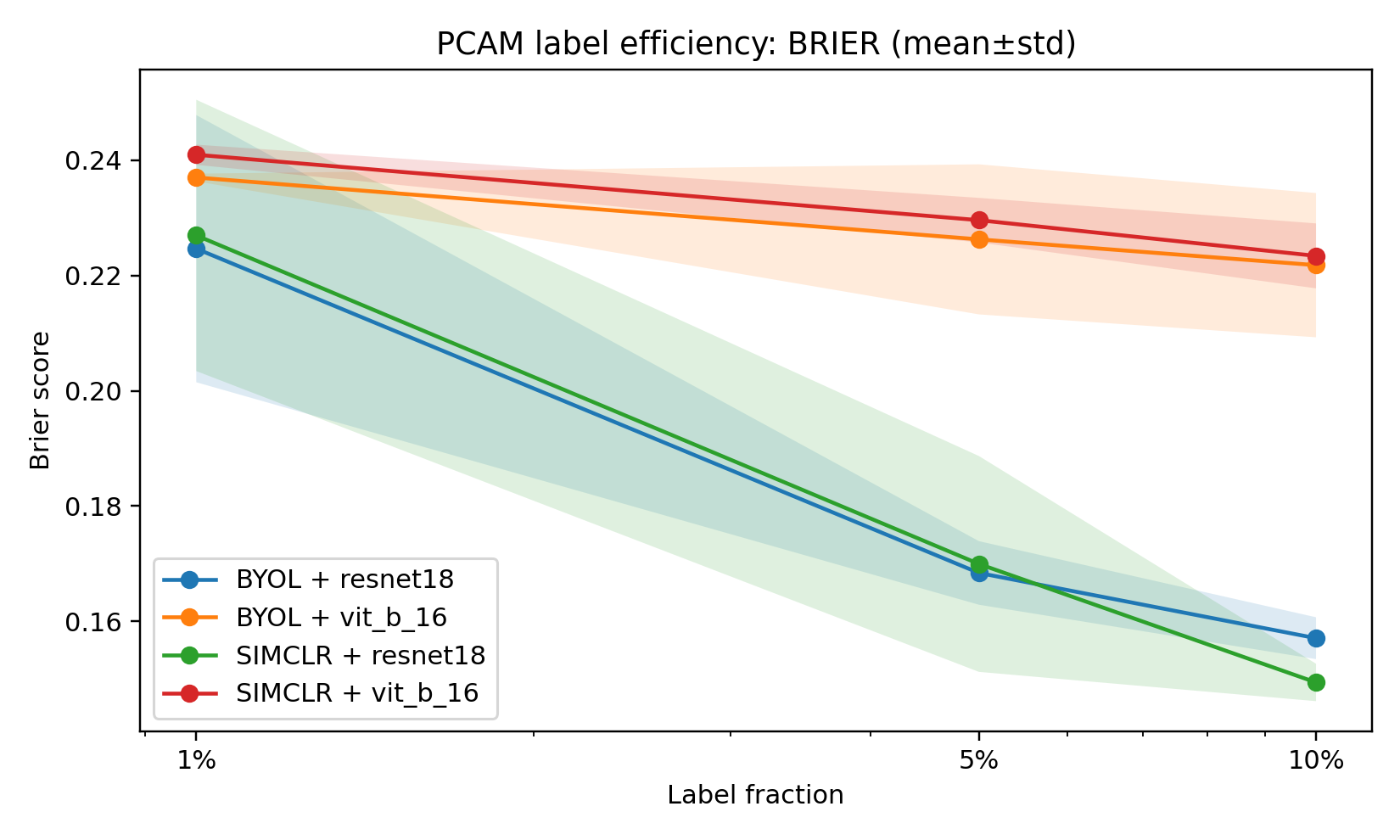}
  \centerline{\small (b) Brier score}
\end{minipage}
\caption{\textbf{PCam calibration and probability-quality diagnostics.} Lower values are better for both ECE and Brier score. ResNet-18 SimCLR/BYOL improve markedly as the label fraction increases.\label{fig:calibration}}
\end{center}
\vspace{-3mm}
\end{figure}

Ranking and confidence reliability need not move together: AUROC improvements may not translate to reliable confidence. We therefore report ECE and Brier alongside AUROC/accuracy/F1. \autoref{fig:calibration} shows that calibration generally improves with more labels on PCam, especially for ResNet-18 SimCLR and BYOL.

\subsection{Federated, MAE, OOD, and transfer diagnostics}

\begin{table}[H]
\footnotesize
\setlength{\tabcolsep}{2pt}
\renewcommand{\arraystretch}{0.82}
\centering
\begin{tabularx}{0.98\textwidth}{lXcc}
\toprule
\textbf{Study} & \textbf{Setting} & \textbf{AUROC} & \textbf{ECE / note} \\
\midrule
\rowcolor{LightBlue} Federated SSL & PCam 1\%, centralized SimCLR & $0.687\pm0.082$ & ECE $0.101\pm0.006$ \\
Federated SSL & PCam 1\%, federated SimCLR & $0.860\pm0.003$ & ECE $0.004\pm0.001$ \\
\rowcolor{LightBlue} Federated SSL & PCam 10\%, centralized SimCLR & $0.869\pm0.004$ & ECE $0.086\pm0.008$ \\
Federated SSL & PCam 10\%, federated SimCLR & $0.874\pm0.001$ & ECE $0.006\pm0.002$ \\
\rowcolor{LightBlue} MAE baseline & PCam 1\%, MAE-inspired & $0.708\pm0.006$ & ECE $0.089\pm0.018$ \\
OOD diagnostic & PANDA$\to$PCam 5\% & raw $0.031\pm0.025$ & flip $0.969\pm0.025$ \\
\rowcolor{LightBlue} OOD diagnostic & PCam$\to$PANDA 5\% & raw $0.243\pm0.274$ & flip $0.757\pm0.274$ \\
Transfer & PCam$\to$colorectal 10\% & $0.942\pm0.006$ & F1 $0.686\pm0.015$ \\
\bottomrule
\end{tabularx}
\caption{\textbf{Diagnostic and extension results.} Compact summary of the main non-benchmark evidence.\label{tab:extensions}}
\vspace{0.8mm}
\end{table}
\FloatBarrier

\autoref{tab:extensions} summarizes the main extension studies. In the PCam diagnostic setting, federated SimCLR achieved competitive AUROC and substantially lower ECE than the split-matched centralized baseline. Because this experiment uses a compact simulation with three clients and five communication rounds, we interpret the result as calibration-oriented diagnostic evidence rather than as a general claim that federated SSL universally outperforms centralized SSL. The MAE-inspired baseline is competitive at 1\% labels but shows different calibration behavior, so ranking performance and confidence reliability should be interpreted separately.

\begin{figure}[!htbp]
\begin{center}
\includegraphics[width=0.94\textwidth]{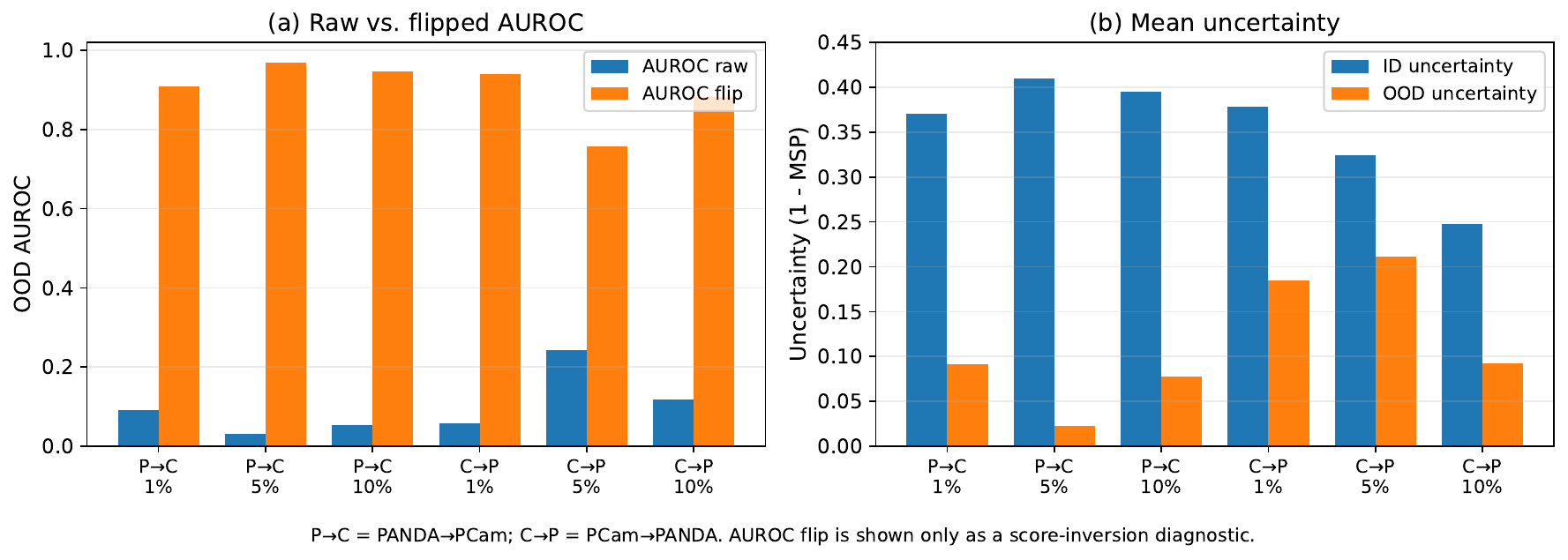}
\caption{\textbf{OOD diagnostics between PCam and PANDA.} AUROC$_{\mathrm{flip}}$ is shown only to diagnose score inversion. Lower OOD uncertainty than ID uncertainty indicates overconfident OOD predictions.\label{fig:ood}}
\end{center}
\vspace{-3mm}
\end{figure}

\autoref{fig:ood} reports OOD AUROC using an uncertainty score derived from the probe's confidence. We observe marked directionality: AUROC$_{\mathrm{raw}}<0.5$ indicates score inversion, meaning the model is more confident on OOD than ID samples. AUROC$_{\mathrm{flip}}$ is therefore only an inversion diagnostic, not a claim of successful OOD detection. This behavior is clinically important because a model that is highly confident on shifted inputs may appear reliable under conventional accuracy metrics while failing as a confidence-aware screening tool. A likely cause is that frozen SSL features optimize augmentation invariance rather than class separation, so MSP confidence (already known to degrade under shift) need not track familiarity, an effect accentuated by label-free pretraining rather than unique to it.

\subsection{Representation-space evidence}

\begin{figure}[H]
\begin{center}
\includegraphics[width=0.68\textwidth]{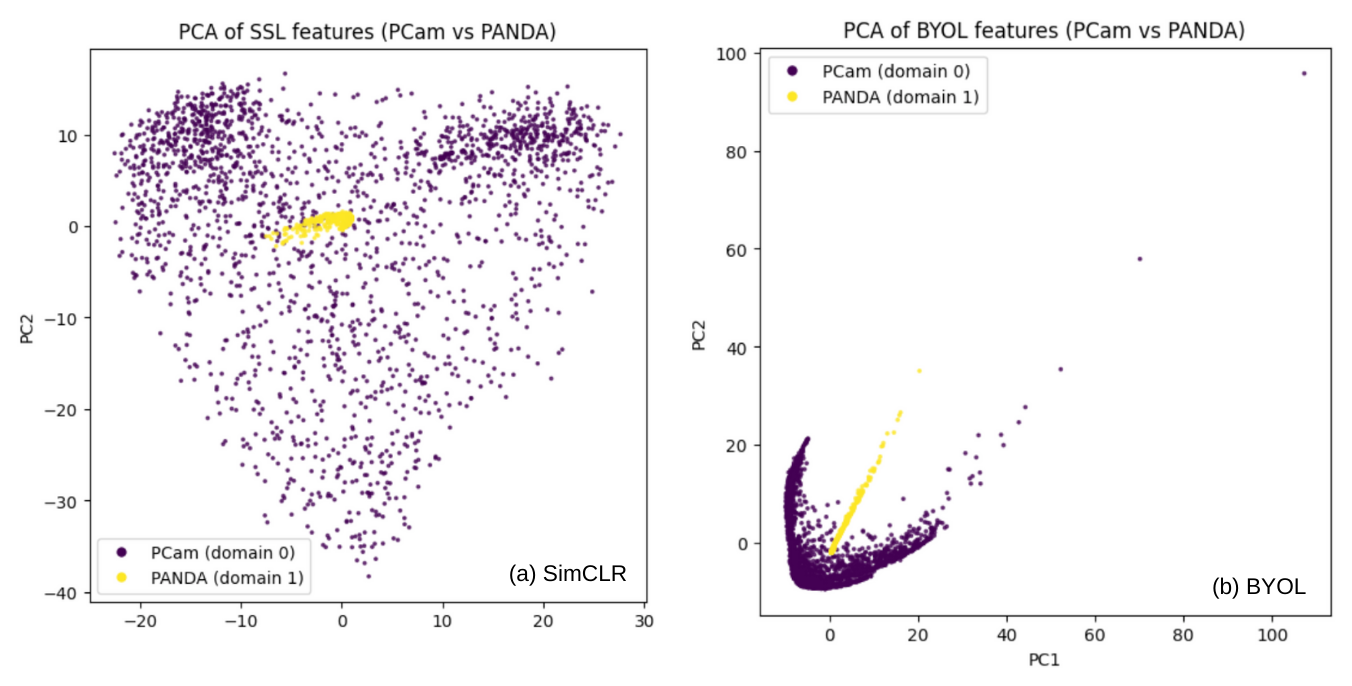}
\caption{\textbf{PCA projections of SSL embeddings} for PCam and PANDA. Left: SimCLR; right: BYOL. PANDA samples form a compact, distinct region, consistent with domain-shift behavior.\label{fig:pca}}
\end{center}
\vspace{-3mm}
\end{figure}

\autoref{fig:pca} gives a qualitative view of the learned feature spaces. SimCLR places PANDA samples in a compact region within a broader PCam cloud, while BYOL produces a different curved geometry. The projection supports that PCam and PANDA are not interchangeable domains and that confidence scores can behave differently by direction.

\FloatBarrier

\section{Conclusion}
\label{sec:conclusion}

We presented a reliability-aware benchmark of self-supervised learning for low-label histopathology classification. Across PCam and PANDA, SSL improves label efficiency, but the strength and stability of these gains depend strongly on dataset construction, supervision quality, and domain shift. PCam offers a more controlled setting where SSL benefits are clearer, while PANDA exposes the limitations of patch-level proxies derived from slide-level annotations. Beyond AUROC, calibration and OOD diagnostics reveal failure modes that performance-only reporting would miss. These findings support SSL for annotation-limited pathology research, but biomedical SSL benchmarks should include reliability, transparency, and robustness checks before clinical interpretation. For practitioners, we recommend shared seeded splits across methods, mean$\pm$std over split$\times$seed runs, a calibration metric paired with every ranking metric, and a cross-dataset confidence check to expose overconfidence under shift.

\textbf{Limitations.} This is a controlled patch-level benchmark, not a clinical deployment study. PANDA labels are derived from slide-level information and may be weakly informative or noisy at the patch level; ViT-B/16 uses one fixed split with three seeds due to compute constraints; the federated study is a compact diagnostic (three clients, five rounds); and whole-slide aggregation is not evaluated. The method scope (SimCLR, BYOL, MAE-inspired) also leaves other SSL families (MoCo, Barlow Twins, DINOv2) and larger backbones (ResNet-50, ConvNeXt) for future work, alongside multi-center WSI-level validation and prospective clinical reliability testing.

\vspace{-1mm}
\section*{Conflict of interests}
\label{sec:CONFLICT-OF-INTERESTS}
The authors declare no competing interests.

\vspace{-1mm}
\section*{Funding}
\label{sec:FUNDING}
This research received no specific grant from any funding agency in the public, commercial, or not-for-profit sectors.

\vspace{-1mm}
\section*{Availability of data and software code}
\label{sec:AVAILABILITY}
The datasets used in this study are publicly available: PatchCamelyon, PANDA, and the colorectal histology collection. The source code supporting the experiments is publicly available at \url{https://github.com/alicemkyn/ssl-histopath-reliability}.

\vspace{-1mm}
\bibliographystyle{unsrt}
\bibliography{references}

\end{document}